\documentclass[runningheads]{llncs}

\usepackage{eccv}

\usepackage{eccvabbrv}

\usepackage{graphicx}
\usepackage{booktabs}

\usepackage[accsupp]{axessibility}  

\usepackage{booktabs}
\usepackage{multirow}
\usepackage{amssymb}
\usepackage{amstext}
\usepackage{graphicx}
\usepackage{makecell}
\usepackage{wrapfig}

\usepackage{caption}
\usepackage{subcaption}
\usepackage{enumitem}

\usepackage{hyperref}

\usepackage{orcidlink}

\begin{document}

\title{Learning to Adapt SAM for Segmenting Cross-domain Point Clouds} 


\author{Xidong Peng\inst{1} \and Runnan Chen\inst{2} \and Feng Qiao\inst{3} \and Lingdong Kong\inst{4} \and Youquan Liu\inst{5} \and Yujing Sun\inst{2} \and Tai Wang\inst{6} \and Xinge Zhu\inst{7\star} \and Yuexin Ma\inst{1}\thanks{Corresponding author. This work was supported by NSFC (No.62206173), Natural Science Foundation of Shanghai (No.22dz1201900), Shanghai Sailing Program (No.22YF1428700), MoE Key Laboratory of Intelligent Perception and Human-Machine Collaboration (ShanghaiTech University), Shanghai Frontiers Science Center of Human-centered Artificial Intelligence (ShangHAI).}}

\authorrunning{Peng. et al.}


\institute{$^1$ShanghaiTech University, $^2$The University of Hong Kong, \\ $^3$RWTH Aachen University, $^4$National University of Singapore, \\ $^5$Hochschule Bremerhaven, $^6$Shanghai AI Laboratory, \\ $^7$The Chinese University of Hong Kong \\ \email{\{pengxd,mayuexin\}@shanghaitech.edu.cn}}

\maketitle

\begin{abstract}
Unsupervised domain adaptation (UDA) in 3D segmentation tasks presents a formidable challenge, primarily stemming from the sparse and unordered nature of point clouds. Especially for LiDAR point clouds, the domain discrepancy becomes obvious across varying capture scenes, fluctuating weather conditions, and the diverse array of LiDAR devices in use. Inspired by the remarkable generalization capabilities exhibited by the vision foundation model, SAM, in the realm of image segmentation, our approach leverages the wealth of general knowledge embedded within SAM to unify feature representations across diverse 3D domains and further solves the 3D domain adaptation problem. Specifically, we harness the corresponding images associated with point clouds to facilitate knowledge transfer and propose an innovative hybrid feature augmentation methodology, which enhances the alignment between the 3D feature space and SAM's feature space, operating at both the scene and instance levels. Our method is evaluated on many widely-recognized datasets and achieves state-of-the-art performance.
  \keywords{Unsupervised Domain Adaptation \and 3D Segmentation \and Feature Alignment \and Vision Foundation Model}
\end{abstract}

\section{Introduction}
\begin{figure}[t]
    \centering
    \includegraphics[width=1.0\columnwidth]{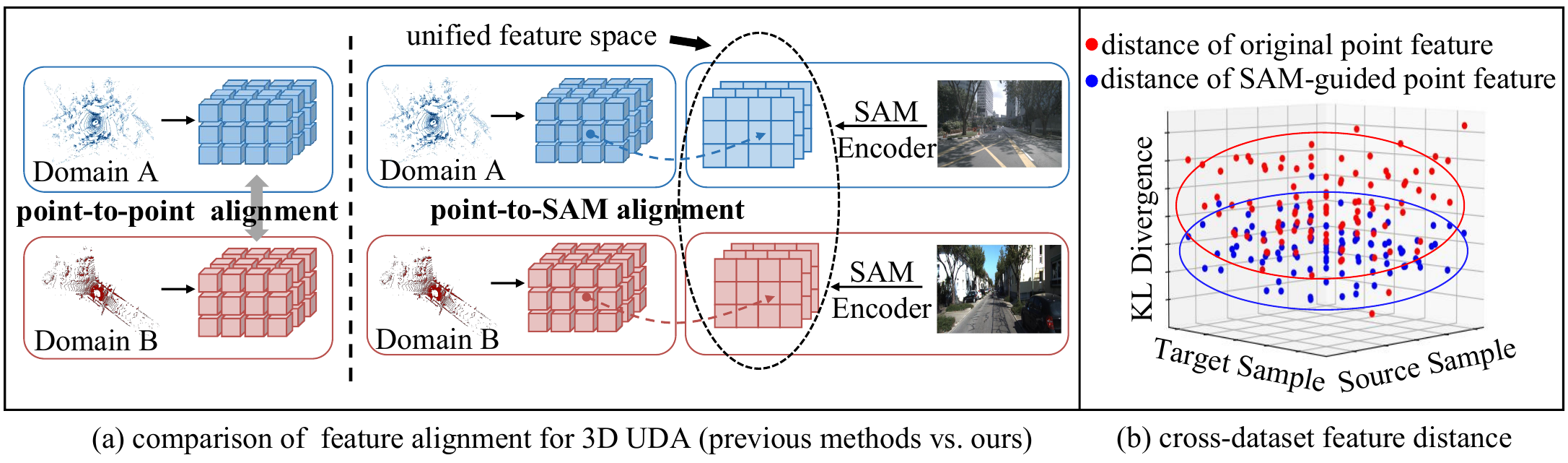}
    \vspace{-6mm}
    \caption{(a) Comparison of 3D UDA paradigms. Different from aligning two point feature domains directly, our method makes both the source domain and target domain align with the SAM feature space. (b) Visualization of the feature distance across different datasets, where smaller values indicate a more similar distribution. It is obvious that after mapping to SAM feature space, point feature distributions from disparate domains become much more aligned. 
    }
    \label{fig_teaser}
    \vspace{-6mm}
\end{figure}

3D scene understanding is fundamental for many real-world applications, such as autonomous driving, robotics, smart cities, etc. Based on the point cloud, 3D segmentation is a critical task for scene understanding, which requires assigning semantic labels for each point. Current deep learning-based solutions~\cite{zhu2021cylindrical,xu2023human} rely heavily on massive annotated data, which are high-cost and lack generalization capability for handling domain shifts. Unsupervised domain adaptation is significant for alleviating data dependency. However, unlike images with dense and regular representation, point clouds, especially LiDAR point clouds of large scenes, are unstructured and sparse, and have overt differences in patterns for various capture devices. Although some studies~\cite{yi2021complete, saltori2022cosmix, Shaban2023LiDARUDAST} have extended 2D techniques to solve the 3D UDA problem, the performance is still limited due to the essential defect of point cloud representation.

Considering that RGB cameras yield dense, color-rich, and structured data, and more importantly, they represent minor discrepancies across various devices, certain 3D UDA methods~\cite{jaritz2020xmuda,cardace2023exploiting, cao2023mopa} utilize the synergy of LiDAR and camera capabilities to achieve more comprehensive and precise perception, and further enhance adaptation capabilities for 3D segmentation tasks. However, these methods usually train 2D and 3D networks simultaneously, demanding substantial online computing resources. Vision foundation models (VFMs), such as the Segment Anything Model (SAM)~\cite{kirillov2023segment}, have garnered significant attention due to their remarkable performance in addressing open-world vision tasks. Such models are trained on massive image data with tremendous parameters. Compared with a common model trained on limited data, VFMs have more general knowledge and much stronger generalization capability. Many works such as ~\cite{chen2023clip2scene,chen2023towards} have emerged recently to transfer the general 2D vision knowledge of VFMs to 3D and have achieved promising performance.

Based on SAM, focusing on image segmentation, we propose a novel paradigm for 3D UDA segmentation. As shown in Fig.~\ref{fig_teaser}(a), different from previous UDA approaches that strive to align the target domain to the source domain so that the model trained on labeled source data can also work on target data without annotation, our method makes both the source domain and target domain align with the SAM feature space. SAM feature space contains more general knowledge, which provides a friendly space to unify the feature representation from different domains. We utilize RGB images to assist point clouds in our framework. However, unlike the methods mentioned above only using images to provide auxiliary information, we take images as a bridge to align diverse 3D feature spaces to the SAM feature space, so we do not need to train extra 2D networks and we can process the image offline for less computing resources. 
Moreover, considering that the 3D feature space created by the source-domain data and the target-domain data is still much smaller than the SAM feature space, we propose a hybrid feature augmentation method at both scene and instance levels to generate more 3D data with diverse feature patterns in a broader data domain, which can further benefit the 3D-to-SAM feature alignment. In particular, we make full use of the masks generated by SAM to mix instance-level point clouds with the other domains. This technique can maintain the geometric completeness of instances, which is beneficial for semantic recognition.

To verify that our idea of SAM-guided UDA is reliable, we randomly choose data from the source and target datasets and calculate the differences of feature distributions from source and target domain by KL divergence. The qualitative results are shown in Fig.~\ref{fig_teaser}(b), where the feature distribution differences after mapping to SAM feature space truly become much smaller. Moreover, to verify the effectiveness of our method, we compare it with current SOTA works on extensive 3D UDA segmentation settings and our method outperforms others by a large margin, improving about $14\%$ mIoU for VirtualKITTI-to-SematicKITTI, about $15\%$ mIoU for Waymo-to-nuScenes, and about $20\%$ mIoU for nuScenes-to-SemanticKITTI domain adaptation. Surprisingly, our unsupervised method achieves comparable performance with the supervised method for city-changing and light-changing settings on the nuScenes dataset. We also test our method on more challenging tasks, such as panoptic segmentation and domain generalization, showing that our method is robust and has good generalization capability.

In summary, our contributions are as follows:

\begin{itemize}[leftmargin=0.5cm, itemindent=0cm]
    \item We propose a novel unsupervised domain adaptation approach for 3D segmentation, leveraging the foundational model SAM to guide the alignment of features from diverse 3D data domains into a unified domain.
    \item We introduce a hybrid feature augmentation strategy at both scene and instance levels, generating more distinct feature patterns across a broader data domain for better feature alignment.
    \item We conduct extensive experiments on large-scale datasets and achieve SOTA performance.
\end{itemize}

\section{Related Work}
\subsection{Point Cloud Semantic Segmentation}

Point cloud semantic segmentation~\cite{zhu2021cylindrical,guo2020deep} is a rapidly evolving field, and numerous research works have contributed to advancements in this area. The pioneering approach PointNet~\cite{qi2017pointnet} directly processes point clouds without voxelization and revolutionizes 3D segmentation by providing a novel perspective on point cloud analysis. Further, PointNet++~\cite{qi2017pointnet++} extends PointNet with hierarchical feature learning through partitioning point clouds into local regions. To handle sparse point cloud data efficiently within large-scale scenes, a framework called SparseConvNet~\cite{graham20183d} has been specifically crafted. It excels in processing sparse 3D data and has been effectively utilized in various applications, including 3D semantic segmentation. MinkUNet~\cite{choy20194d} represents a significant advancement in point cloud semantic segmentation. Employing multi-scale interaction networks, MinkUNet enhances the segmentation of point clouds, effectively addressing the challenges posed by 3D spatial data. Our 3D segmentation networks are the popular SparseConvNet and MinkUNet. Due to the sparse characteristics of point cloud data, many current methods~\cite{yan20222dpass, krispel2020fuseseg, he2022multimodal} add corresponding dense image information to facilitate point cloud segmentation tasks. Our method also uses image features to assist point cloud segmentation, and additionally, we utilize the 2D segmentation foundation model to achieve effective knowledge transfer.

\subsection{Domain Adaptation for 3D Segmentation}

Unsupervised Domain Adaptation (UDA) aims at transferring knowledge learned from a source annotated domain to a target unlabelled domain, and there are already several UDA methods proposed for 2D segmentation~\cite{chang2019all,zhang2020joint, kim2020learning, zou2018unsupervised}. In recent years, domain adaptation techniques have gained increasing traction in the context of 3D segmentation tasks. \cite{yi2021complete} leverage a "Complete and Label" strategy to enhance semantic segmentation of LiDAR point clouds by recovering underlying surfaces and facilitating the transfer of semantic labels across varying LiDAR sensor domains. CosMix~\cite{saltori2022cosmix} introduces a sample mixing approach for UDA in 3D segmentation, which stands as the pioneering UDA approach utilizing sample mixing to alleviate domain shift. It generates two new intermediate domains of composite point clouds through a novel mixing strategy applied at the input level, mitigating domain discrepancies. However, due to the sparsity and irregularity of the point cloud, the disparity across different point cloud data domains is larger compared to that across 2D image domains, which makes it difficult to mitigate the variation across domains.

With the development of multi-modal perception~\cite{bai2022transfusion,cong2023weakly} in autonomous driving, prevalent 3D datasets~\cite{fong2022panoptic,mei2022waymo,behley2019semantickitti,geyer2020a2d2} include both 3D point clouds and corresponding 2D images, making leveraging multi-modality for addressing domain shift challenges in point clouds convenient. xMUDA~\cite{jaritz2020xmuda, jaritz2022cross} shows the power of combining 2D and 3D networks within a single framework, which achieves outstanding performance by aggregating the scores from these two branches. This achievement is attributed to the complementary nature resulting from the diverse modalities processed by each branch. \cite{peng2021sparse} introduce Dynamic Sparse-to-Dense Cross-Modal Learning (DsCML) to enhance the interaction of multi-modality information, ultimately boosting domain adaptation sufficiency, while \cite{cardace2023exploiting} elucidate this complementarity of image and point cloud through an intuitive explanation centered on the effective receptive field, and proposes to feed both modalities to both branches. However, in practice, training two networks with distinct architectures is difficult to converge and demands substantial computing resources due to increased memory. Our method uses the pre-trained foundation model to process the image data, guaranteeing the quality of the image features and enabling the training process to focus on the 3D model.

\subsection{Vision Foundation Models}

The rise of foundation models~\cite{devlin2018bert,jia2021scaling,touvron2023llama} has garnered significant attention which are trained on extensive datasets, consequently demonstrating exceptional performance. Foundation models~\cite{zou2023segment,wang2023seggpt} have seen significant advancements in the realm of 2D vision, and several research studies extend these foundation models to comprehend 3D information. Representative works CLIP~\cite{radford2021learning} leverage contrastive learning techniques to train both text and image encoders. CLIP2Scene~\cite{chen2023clip2scene} extends the capabilities of CLIP by incorporating a 2D-3D calibration matrix to facilitate a deeper comprehension of 3D scenes and OpenScene\cite{peng2023openscene} focuses on zero-shot learning for 3D scenes through aligning point features in CLIP feature space to enable open vocabulary queries for 3D points. The Meta Research team recently launched the `Segment Anything Model'~\cite{kirillov2023segment}, trained on an extensive dataset of over 1 billion masks from 11 million images. Utilizing efficient prompting, SAM can generate high-quality masks for image instance segmentation. The integration of flexible prompting and ambiguity awareness enables SAM with robust generalization capabilities for various downstream segmentation challenges. Many methods~\cite{chen2023towards,chen2023clip2scene,liu2023segment} take it as an off-the-shelf tool and distillate the knowledge to solve 3D problems by 2D-3D feature alignment. In our work of tackling the UDA of 3D segmentation, we utilize SAM to provide 2D prior knowledge for 3D feature alignment in a wider data domain.

\section{Method}
\subsection{Problem Statement}

We explore UDA for 3D segmentation, in which we have the source domain, denoted as $D_S = \{P_S, I_S, Y_S\}$ with paired input, namely point cloud $P_S$ and image $I_S$, as well as annotated labels $Y_S$ for each point, and the target domain denoted as $D_T = \{P_T, I_T\}$ without any annotation. Using these data, we train a 3D segmentation model that can generalize well to the target domain. 3D data from various domains have obvious differences in distribution and patterns, leading to over-fitting problems when models trained in one domain try to analyze data from another. The main solution is to align different features despite domain differences to achieve the generalization capability of the model. 
Different from previous works, our novel paradigm is to map data from distinct domains into a unified feature space, ensuring the model performs consistently across domains.

\begin{figure*}[h]
    \centering
    \includegraphics[width=0.95\columnwidth]{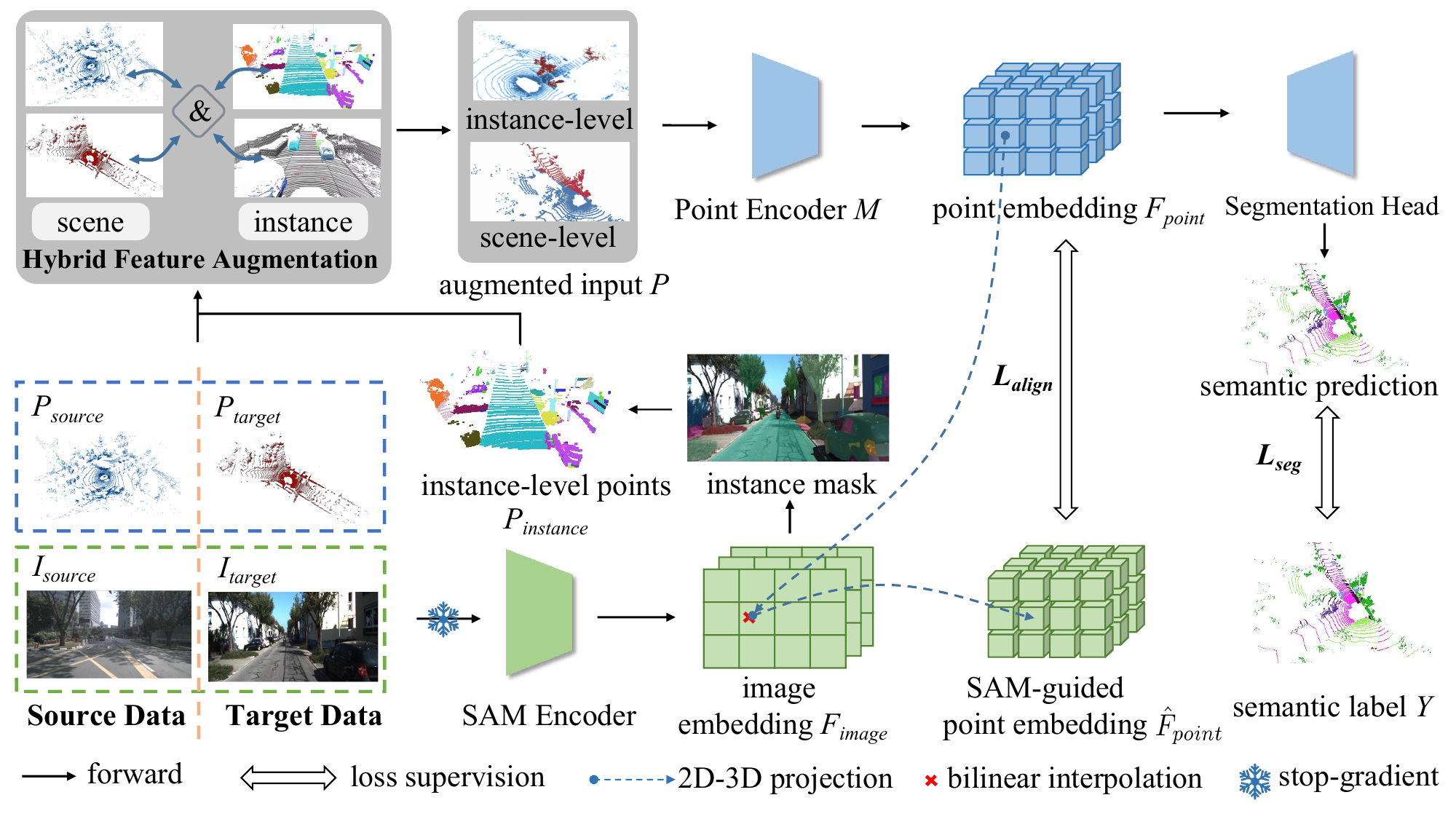}
    \vspace{-3mm}
    \caption{Pipeline of our method. The point cloud is fed into the point encoder for point embeddings at the top, and the corresponding images are passed through the SAM encoder for image embeddings at the bottom, from which we obtain SAM-guided point embedding with the 2D-3D projection. Alignment loss $L_{align}$ is calculated based on the SAM-guided features and original features. Furthermore, augmented inputs provide diverse feature patterns boosting the 3D-to-SAM feature alignment.}
    \label{fig_framework}
    \vspace{-6mm}
\end{figure*}

\subsection{Framework Overview}

The vision foundation model, SAM, is trained by massive image data, which contains general vision knowledge and provides a friendly feature space to unify diverse feature representations. Taking 2D images as the bridge, the 3D feature space of different domains can be indirectly unified by bringing them closer to the SAM feature space based on 2D-to-3D knowledge distillation. Based on this, we design a novel SAM-guided UDA method for 3D segmentation, as Fig.~\ref{fig_framework} shows. Specifically, given a point cloud input $P$, the point encoder $M$ generates a point embedding $F_{point} \in \mathbb{R}^{n \times d}$ in the $d$-dimensional latent feature space. Concurrently, the corresponding image input $I$ is passed through the SAM encoder for a c-channels image embedding $F_{image} \in \mathbb{R}^{h \times w \times c}$. Utilizing the correspondence between the point cloud and image, we acquire SAM-guided point embedding $\hat{F}_{point} \in \mathbb{R}^{n \times d}$ to compute the alignment loss $L_{align}$ with the original point embedding $F_{point}$, serving the purpose of using SAM as a bridge to integrate the features of diverse data domains into a unified feature space. Notably, during training, the input for feature alignment consists of data from both source and target domains. We named this process as \textbf{SAM-guided Feature Alignment}. At the same time, as for labeled data $Y$, segmentation loss $L_{seg}$ is also calculated as semantic supervision. During model training, only the point cloud branch of the whole pipeline is trained, and the gradient is not calculated in the image branch, which makes our method more lightweight. Furthermore, a \textbf{Scene-Instance Hybrid Feature Augmentation} is designed, which consists of scene-level and instance-level mix-up strategy. These mix-up strategies boost the variance of training data and generalize the network capability under the convex combination of the source domain and target domain data. Notably, the instance-level feature augmentation could maintain the local geometric relationship between two domains and make the subsequent alignment efficient.

\subsection{SAM-guided 3D Feature Alignment}

Previous UDA methods usually align the feature space of the target domain to that of the source domain so that the model trained on the source domain with labeled data can also recognize the data from the novel domain. However, the distributions and patterns of 3D point clouds in various datasets have substantial differences, making the alignment very difficult. SAM~\cite{kirillov2023segment}, a 2D foundation model, is trained with a huge dataset of 11M images, granting it robust generalization capabilities to address downstream segmentation challenges effectively. If we can align features extracted from various data domains into the unified feature space represented by SAM, the model trained on the source domain can effectively handle the target data with the assistance of the universal vision knowledge existing in the SAM feature space.

We focus on training a point-based 3D segmentation model, while SAM is a foundation model trained on 2D images, which presents a fundamental challenge: how to bridge the semantic information captured in 2D images with the features extracted from 3D points. Most outdoor large-scale datasets with point clouds and images provide calibration information to project the 3D points into the corresponding images. With this information, we can easily translate the coordination of points $P$ from the 3D LiDAR coordinate system $P_{lidar}$ to the 2D image coordinate system $P_{image}$. This transformation can be formally expressed as  Eq.~\ref{eq_proj}, where rotation $R_{ext}$ and translation $T_{ext}$ represent the extrinsic parameters of the camera, and matrix $K$ represents the intrinsic parameters of the camera.

\begin{equation}
\label{eq_proj}
zP_{image} = K(R_{ext} P_{lidar} + T_{ext})
\end{equation}

Once we calculate the projected 2D positions of points in the image coordinate system, we can determine their corresponding positions in the SAM-guided image embedding $F_{image}$, which is generated from the image by the SAM feature extractor. As the positions of points in the image embedding typically are not integer values, we perform bilinear interpolation based on the surrounding semantic features in the image embedding corresponding to the point, which to some extent alleviates the effect of calibration errors and allows us to derive the SAM-guided feature of each point, denoted as Eq.~\ref{eq_biliner}.

\begin{equation}
\label{eq_biliner}
\hat{F}_{point} = \textbf{Bilinear}(F_{image}, \ P_{image})
\end{equation}

Then, the original point embedding $F_{point}$ from both source and target domains are all required to align with their corresponding SAM-guided features $\hat{F}_{point}$. Specifically, we utilize the cosine function to measure the similarity of $F_{point}$ and $\hat{F}_{point}$, employing it as the alignment loss $L_{align}$ during training. With the supervision of $L_{align}$, features obtained by the point encoder $M$ will gradually converge towards the feature space represented by SAM, achieving the purpose of extracting features within a unified feature space from the input of different domains. The formulation of the loss function for feature alignment is shown as Eq.~\ref{eq_loss}.

\begin{equation}
\label{eq_loss}
L_{align} = 1 - \textbf{cos}(\hat{F}_{point}, \ F_{point})
\end{equation}

\begin{figure*}[h]
    \centering
    \includegraphics[width=1.0\columnwidth]{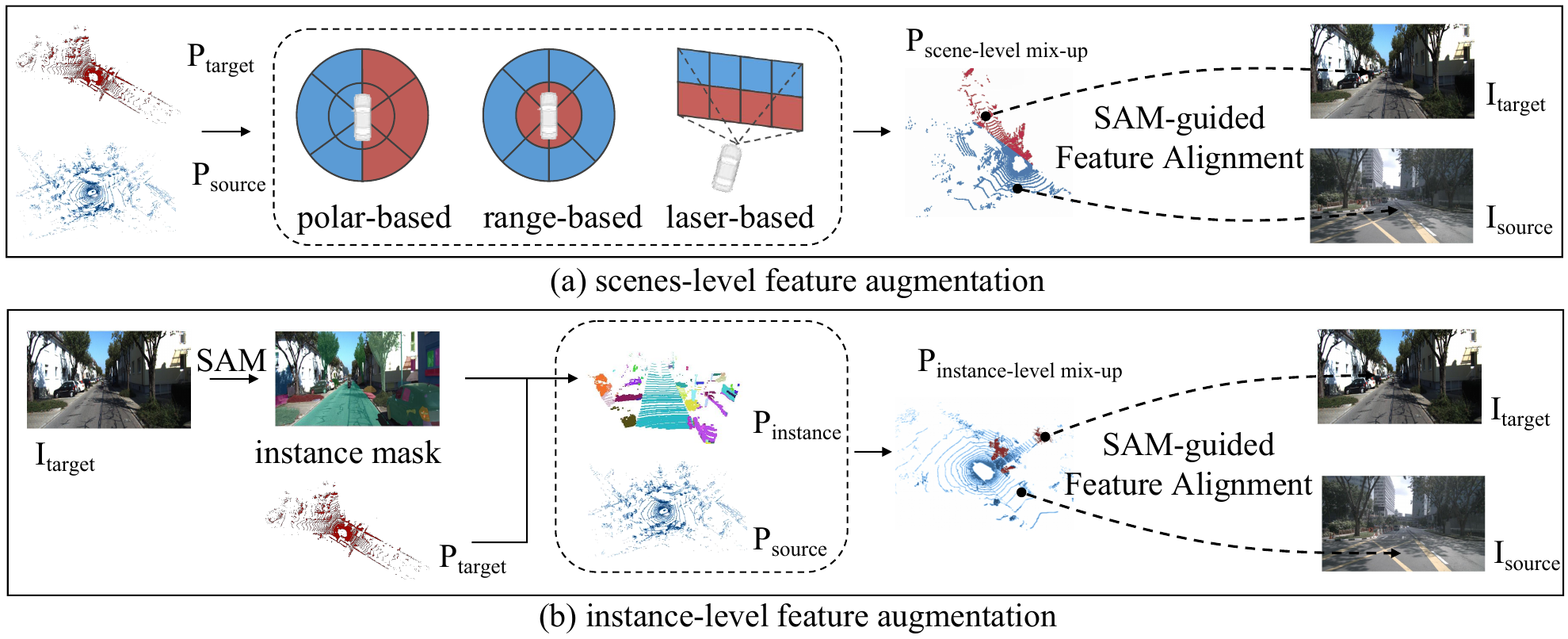}
    \vspace{-6mm}
    \caption{Hybrid feature augmentation by data mixing for better 3D-to-SAM feature alignment. Part(a) illustrate all the scene-level approaches including polar-based, range-based, and laser-based point mix-up, where different color represents points from distinct domain. Part(b) shows the data flow of mixing the point data with instance-level data from another domain with an instance mask, where we take source data as an example for instance-level point generation and vice versa. 
    }
    \label{fig_aug}
      \vspace{-6mm}
\end{figure*}

\subsection{Scene-Instance Hybrid Feature Augmentation}

3D point features of the source-domain data and the target-domain data only cover subsets of the 3D feature space, which are limited to align with the whole SAM feature space with more universal knowledge. Therefore, more 3D data with diverse feature patterns in a broader data domain is needed to achieve more effective 3D-to-SAM feature alignment. 

Previous works~\cite{xiao2022polarmix,kong2023lasermix} usually focus on synthesizing data by combining data in the source domain and target domain at the scene level, including polar-based, range-based, and laser-based, as shown in Fig.\ref{fig_aug}(a). Polar-based point mix-up selects semi-circular point cloud data from two different domains based on the polar coordinates of the point cloud. Range-based point mix-up divides the point cloud by its distance from the center, synthesizing circular point data close to the center and ring point data farther away from the center. Laser-based point mix-up determines the part of point clouds based on the number of laser beams, combining points with positive and negative laser pitch angles from different domains for synthesis. These ways of scene-level feature augmentation can maintain the general pattern of LiDAR point clouds as much as possible and improve the data diversity. Moreover, they are simple to process without any requirement for additional annotations such as real or pseudo-semantic labels. We adopt these three kinds of scene-level data augmentation in our method.

However, scene-level data augmentation will, to some extent, destroy the completeness of the point cloud of instances in the stitching areas and affect the exploitation of local geometric characteristics of point clouds. To further increase the data diversity and meanwhile keep the instance feature patterns of LiDAR point clouds for better semantic recognition, we propose an instance-level augmentation method. Benefiting from the instance mask output from SAM, we can thoroughly exploit the instance-level geometric features. Compared with pre-trained 3D segmentation models, SAM provides more accurate and robust instance masks and enables us to avoid extra warm-up for a pre-train model, simplifying the whole training process. Therefore, we perform instance-level data synthesis as Fig.\ref{fig_aug}(b) shows. Specifically, we begin by employing SAM to generate instance masks for input images from either the source or target domain (We take target data as the example in the figure). Next, we use the calibration matrix to project the corresponding point cloud into the image. The instance information of each point is determined according to whether the projection position of the point cloud falls within a specific instance mask, and then we randomly select some points with 20 $\sim$ 30 specific instances, mixed with the point cloud from the other domain by direct concatenation to achieve point augmentation at the instance level.

In practice, we combine all the ways of feature augmentation at both scene level and instance level with a random-selection strategy for a more comprehensive feature augmentation, which generates a more diverse set of point cloud data with varied feature patterns. Then, the augmented points are fed into the point encoder $M$ to obtain the point embedding $F_{point}$ with distinct feature patterns in a broader data domain beyond the source domain and target domain for more effective SAM-guided feature alignment. Notably, to maintain the consistency of the point cloud and the image, we extract SAM-guided point embedding based on the corresponding original image embedding.



\section{Experiment}
We first introduce datasets and implementation details. After that, we explore several domain shift scenarios and conduct comparisons for 3D segmentation. Then, we conduct extensive ablation studies to assess submodules of our method. Finally, we extend our method to more challenging tasks to show its generalization capability.

\subsection{Dataset Setup}
\label{sec_setup}

We first follow the benchmark introduced in xMUDA~\cite{jaritz2022cross} to evaluate our method, comprehending four domain shift scenarios, including (1) USA-to-Singapore, (2) Day-to-Night, (3) VirtualKITTI-to-SemanticKITTI and (4) A2D2-to-Semantic-KITTI. The first two leverage nuScenes~\cite{caesar2020nuscenes} as their dataset, consisting of 1000 driving scenes in total with 40k annotated point-wise frames. Specifically, the former differs in the layout and infrastructure while the latter exhibits severe illumination changes between the source and the target domain. The third is more challenging since it is the adaptation from synthetic to real data, implemented by adapting from VirtualKITTI~\cite{gaidon2016virtual} to SemanticKITTI~\cite{behley2019semantickitti} while the fourth involves A2D2~\cite{geyer2020a2d2} and SemanticKITTI as different data domains, where the domain discrepancy lies in the distinct density and arrangement of 3D point clouds captured by different devices since the A2D2 is captured by 16-beam LiDAR and the SemanticKITTI uses 64-beam LiDAR. For the above settings, noted that only 3D points visible from the camera are used for training and testing, specifically, only one image and corresponding points for each sample are used for training. 

Since we only use the image combined with SAM as offline assistance for the training of a 3D segmentation network instead of training a new 2D segmentation network, we focus on comparing the performance of the 3D segmentation network and enable model training with the whole point cloud sample because of less computational cost, even if some part of it is not visible in the images. For the part of the point cloud that cannot be covered by the image, alignment loss is not calculated, only segmentation loss is calculated. Thus, we also compare our method with others trained with the whole 360$^\circ$ view of the point cloud, in which three datasets are involved including nuScenes, SemanticKITTI, and Waymo~\cite{mei2022waymo}. In these settings, we use 6 images in nuScenes covering 360$^\circ$ view, 1 image in SemanticKITTI covering 120$^\circ$ view, and 5 images in Waymo covering 252$^\circ$ view. More information is introduced in the Appendix. For metric, We compute the Intersection over the Union per class and report the mean Intersection over the Union (mIoU). 

\subsection{Implementation Details}

We make source and target labels compatible across these experiments. For all benchmarks in prior multi-modal UDA methods, we strictly follow class mapping like xMUDA for a fair comparison, while we map the labels of the dataset in other experiments into 10 segmentation classes in common. Our method is implemented by using the public PyTorch \cite{paszke2019pytorch} repository MMDetection3D \cite{mmdet3d2020} and all the models are trained on a single 24GB GeForce RTX 3090 GPU. To compare fairly, we use SparseConvNet~\cite{graham20183d} with U-Net architecture as the 3D backbone network when following the benchmark introduced in xMUDA to evaluate our method and use MinkUNet32~\cite{choy20194d} as the 3D backbone network when following the setup of taking the whole 360$^\circ$ point cloud as input, which is also the backbone of the state-of-the-art uni-modal method CosMix. For the image branch, the ViT-h variant SAM model is utilized to generate image embedding for SAM-guided feature alignment and instance masks for hybrid feature augmentation in an offline manner. We keep the proportion of mixed data and normal data from the source and target domain the same during model training. Before the data is fed into the 3D network, data augmentation such as vertical axis flipping, random scaling, and random 3D rotations are widely used like all the compared methods. For the model training strategies, we choose a batch size of 8 for both source data and target data, then mix the data batch for training at each iteration. Besides, we adopt AdamW as the model optimizer and One Cycle Policy as the learning-rate scheduler.

\begin{table*}[h]
\centering
\footnotesize
\caption{Results under four domain shift scenarios introduced by xMUDA. We report all the 3D network performance of compared multi-modal UDA methods in terms of mIoU. Note that the 3D backbone in these experiments is SparseConvNet~\cite{graham20183d}.}
    \vspace{-2mm}

\resizebox{\linewidth}{!}{
\begin{tabular}{l|p{30pt}<{\centering}|c|p{30pt}<{\centering}|c|p{40pt}<{\centering}|c|p{30pt}<{\centering}|c}
\toprule
\textbf{Method} & \multicolumn{2}{c|}{\textbf{USA $\rightarrow$ Singapore}} & \multicolumn{2}{c|}{\textbf{Day $\rightarrow$ Night}} & \multicolumn{2}{c|}{\textbf{v.KITTI $\rightarrow$ Sem.KITTI}} & \multicolumn{2}{c}{\textbf{A2D2 $\rightarrow$ Sem.KITTI}}
\\\midrule
Source only & $62.8$ & \textcolor{gray}{$+0.0$} & $68.8$ & \textcolor{gray}{$+0.0$} & $42.0$ & \textcolor{gray}{$+0.0$} & $35.9$ & \textcolor{gray}{$+0.0$}
\\ \midrule
xMUDA~\cite{jaritz2022cross} & $63.2$ & $+0.4$ & $69.2$ & $+0.4$ & $46.7$ & $+4.7$ & $46.0$ & $+10.1$
\\
DsCML~\cite{peng2021sparse} & $52.3$ & $-10.5$ & $61.4$ & $-7.4$ & $32.8$ & $-9.2$ & $32.6$ & $-3.3$
\\
MM2D3D~\cite{cardace2023exploiting} & $66.8$ & $+4.0$ & $70.2$ & $+1.4$ & $50.3$ & $+8.3$ & $46.1$ & $+10.2$
\\
\textbf{Ours} & $\mathbf{73.6}$ & $\mathbf{+10.8}$ & $\mathbf{70.5}$ & $\mathbf{+1.7}$ & $\mathbf{64.9}$ & $\mathbf{+22.9}$ & $\mathbf{52.1}$ & $\mathbf{+16.2}$
\\\midrule
Oracle & $76.0$ & $-$ & $69.2$ & $-$ & $78.4$ & $-$ & $71.9$ & $-$
\\\bottomrule
\end{tabular}
}
\label{tab_xmuda}
\vspace{-2mm}
\end{table*}

\begin{table*}[h]
\centering
\footnotesize
\caption{Results under four domain shift scenarios with 360$^\circ$ point cloud, where not all the points are visible in the images. We report the 3D network performance in terms of mIoU. Note that the 3D backbone in these experiments is MinkUNet32~\cite{choy20194d}.}
    \vspace{-2mm}


\resizebox{\linewidth}{!}{
\begin{tabular}{l|p{35pt}<{\centering}|c|p{35pt}<{\centering}|c|p{30pt}<{\centering}|c|p{30pt}<{\centering}|c}
\toprule
\textbf{Method} & \multicolumn{2}{c|}{\textbf{nuScenes $\rightarrow$ Sem.KITTI}} & \multicolumn{2}{c|}{\textbf{Sem.KITTI $\rightarrow$ nuScenes}} & \multicolumn{2}{c|}{\textbf{nuScenes $\rightarrow$ Waymo}} & \multicolumn{2}{c}{\textbf{Waymo $\rightarrow$ nuScenes}}
\\\midrule
Source only & $27.7$ & \textcolor{gray}{$+0.0$} & $28.1$ & \textcolor{gray}{$+0.0$} & $29.4$ & \textcolor{gray}{$+0.0$} & $21.8$ & \textcolor{gray}{$+0.0$}
\\ \midrule
PL~\cite{morerio2017minimal} & $30.0$ & $+2.3$ & $29.0$ & $+0.9$ & $31.9$ & $+2.5$ & $22.3$ & $+0.5$
\\
CosMix~\cite{saltori2023compositional} & $30.6$ & $+2.9$ & $29.7$ & $+1.6$ & $31.5$ & $+2.1$ & $30.0$ & $+8.2$
\\
MM2D3D~\cite{cardace2023exploiting} & $30.4$ & $+2.7$ & $31.9$ & $+3.8$ & $31.3$ & $+1.9$ & $33.5$ & $+11.7$
\\
MM2D3D$^*$ & $32.9$ & $+5.2$ & $33.7$ & $+5.6$ & $34.1$ & $+4.7$ & $37.5$ & $+15.7$
\\
\textbf{Ours} & $\mathbf{48.5}$ & $\mathbf{+20.8}$ & $\mathbf{42.9}$ & $\mathbf{+14.8}$ & $\mathbf{44.9}$ & $\mathbf{+15.5}$ & $\mathbf{48.2}$ & $\mathbf{+26.4}$
\\\midrule
Oracle & $70.3$ & $-$ & $78.3$ & $-$ & $79.9$ & $-$ & $78.3$ & $-$
\\\bottomrule
\end{tabular}
}

\label{tab_cosmix}
\vspace{-2mm}
\end{table*}

\subsection{Experimental Results and Comparison}
Tab.~\ref{tab_xmuda} and Tab.~\ref{tab_cosmix} show the experimental results and performance comparison with previous UDA methods for 3D segmentation under the setup introduced in Sec.~\ref{sec_setup}. Each experiment contains two reference methods in common, a baseline model named \textbf{Source only} trained only on the source domain and an upper-bound model named \textbf{Oracle} trained only on the target data with annotations. Tab.~\ref{tab_xmuda} focuses on four domain shift scenarios introduced by xMUDA~\cite{jaritz2022cross} and comparison with these multi-modal methods based on xMUDA such as DsCML~\cite{peng2021sparse} and MM2D3D~\cite{cardace2023exploiting}. Among them, MM2D3D fully exploits the complementarity of image and point cloud and proposes to feed two modalities to both branches, achieving better performance. Our method outperforms it by $+6.8\%$ (USA $\rightarrow$ Singapore), $+0.3\%$ (Day $\rightarrow$ Night), $+14.6\%$ (v.KITTI $\rightarrow$ Sem.KITTI), $+6.0\%$ (A2D2 $\rightarrow$ Sem.KITTI) respectively, because our method aligns all the features into a unified feature space with the guidance of SAM instead of simply aligning features from image and point cloud in 2D and 3D network. Tab.~\ref{tab_cosmix} focuses on the scenarios where not all the point clouds are visible in the images and we re-implement three methods by their official codes. PL~\cite{morerio2017minimal} uses the prediction from the pre-trained model as pseudo labels for unlabelled data to retrain this model, which is widely used in UDA methods. CosMix~\cite{saltori2023compositional} trains a 3D network with only the utilization of a point cloud, which generates new intermediate domains through a mixing scene-level strategy to mitigate domain discrepancies. MM2D3d is the SOTA multi-modal method as described above, but it needs all the points visible in the image for the best performance. Our method surpasses them by at least $+17.9\%$ (nuScenes $\rightarrow$ Sem.KITTI), $+11.0\%$ (Sem.KITTI $\rightarrow$ nuScenes), $+13.0\%$ (nuScenes $\rightarrow$ Waymo), $+14.7\%$ (Waymo $\rightarrow$ nuScenes) respectively by a large margin, since hybrid feature augmentation can provide more intermediate domains and SAM-guided feature alignment can help map the whole point cloud into the unified feature space. 

\begin{figure*}[t]
    \centering
    \includegraphics[width=1.0\columnwidth]{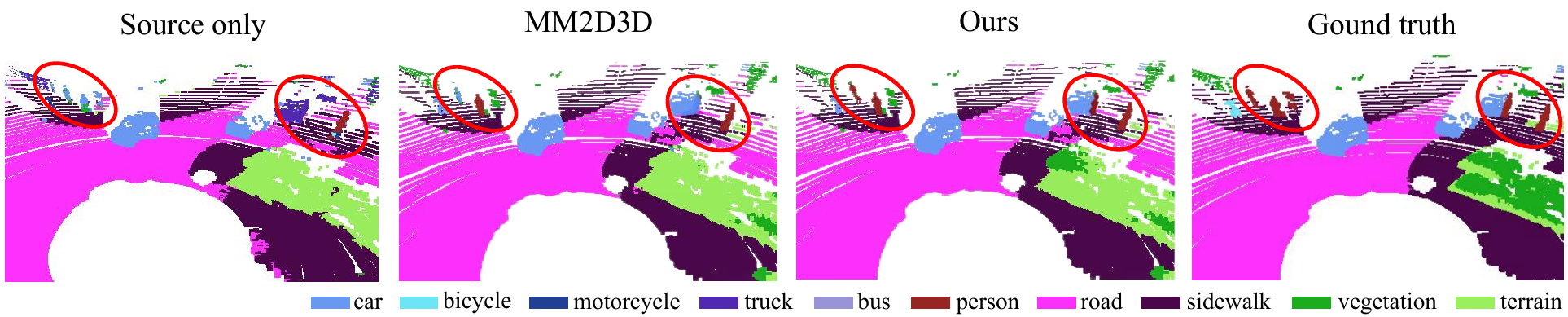}
        \vspace{-2mm}
    \caption{Visualization of the domain adaptation from nuScenes to SemanticKITTI.}
    \label{fig_res}
    \vspace{-2mm}
\end{figure*}

According to the results above, our method outperforms others by a large margin, attributed to the full utilization of the general knowledge provided by SAM. To further prove that the achievement is due to not only SAM but also our novel paradigm, we also use SAM for current SOTA multi-modal UDA work and get results in the row of "MM2D3D*". Because it trains both 2D and 3D networks simultaneously, under its original framework, we can only refine the supervision signals of 2D network using the instance mask output of SAM. As seen from the results, SAM can improve its performance but very limited. Our method not only uses instance masks but also makes full use of the general features extracted by SAM, ensuring the superiority of our method.  Qualitative results are shown in Fig.~\ref{fig_res}, where predictions in the ellipses demonstrate that source-only and MM2D3D models often infer wrong and mingling results, especially for the person category, while our method can provide correct and more fine-grained segmentation. More qualitative results are in the Appendix.



\begin{table}[h]
\centering
\caption{Ablation study. Baseline means the result of the source-only model indicating the lower-bound and Pseudo Label means re-training the model with pseudo labels.}
\vspace{-2mm}
\scriptsize
\resizebox{0.95\linewidth}{!}{
\begin{tabular}{c|ccccc|c}
\toprule
\multirow{2}{*}{\textbf{Setting}} & \multirow{2}{*}{\textbf{Baseline}} & \multirow{2}{*}{\makecell[c]{\textbf{SAM-guided}\\ \textbf{Feature Alignment}}} & \multicolumn{2}{c}{\makecell[c]{\textbf{Hybrid Feature Augmentation}\\ \midrule}} & \multirow{2}{*}{\textbf{Pseudo Label}} & \multirow{2}{*}{\textbf{mIoU}} \\
&                          &                                               & \makecell[c]{\textbf{\quad Scene-level}}              & \makecell[c]{\textbf{Instance-level}}             &                               &                       \\ \midrule
(1) & $\checkmark$ &                   &               &              &  &  $27.7$   \\
(2) & $\checkmark$ &   $\checkmark$    &               &              &  &   $34.0$   \\
(3) & $\checkmark$ &                   &  \makecell[c]{\quad $\checkmark$}     &  $\checkmark$ &  &  $28.6$   \\ \midrule
(4) & $\checkmark$ &   $\checkmark$    &  \makecell[c]{\quad $\checkmark$}     &               &  &  $40.1$    \\
(5) & $\checkmark$ &   $\checkmark$    &                   &  $\checkmark$ &  &  $39.0$   \\
(6) & $\checkmark$ &   $\checkmark$    &  \makecell[c]{\quad $\checkmark$}     &  $\checkmark$ &  &  $44.0$   \\ \midrule
(7) & $\checkmark$ &   $\checkmark$    &  \makecell[c]{\quad $\checkmark$}     &  $\checkmark$ & $\checkmark$ &  $\mathbf{48.5}$   \\
\bottomrule
\end{tabular}
}
\label{tab_abla}
\vspace{-2mm}
\end{table}

\subsection{Ablation Study}
\label{sec_abla}

To show the effectiveness of each module of our method, we conduct ablation studies on nuScenes-to-SemanticKITTI UDA. We also analyze and show the effect of other vision foundation models on our method.

\noindent \textbf{Effectiveness of Model Components} We first analyze the effects of all the submodules in our method in Tab.~\ref{tab_abla}, containing SAM-guided Feature Alignment, Hybrid Feature Augmentation, and Pseudo Label. SAM-guided Feature Alignment aligns all the point features with the corresponding feature embeddings output by SAM, guiding the 3D network map point cloud into the unified feature space represented by SAM while Hybrid Feature Augmentation generates additional point cloud data of the intermediate domain for feature extraction to maximize the effect of feature alignment. Setting (1), (2), (3), and (6) in the table shows that combining the two submodules improves performance by a large margin. Besides, re-training the model with pseudo labels is a strategy widely used in UDA tasks and it also improves the performance.

\noindent \textbf{Effectiveness of Hybrid Feature Augmentation} For the detailed ablation of feature alignment, we adopt hybrid strategies for diverse data with distinct feature patterns, which not only mix up points at the scene level in polar-based, range-based, and laser-based ways but also at the instance level with the help of instance mask output by SAM. Random selection in all these point mix-up ways forms this feature augmentation. Setting (4), (5), (6) in Tab.~\ref{tab_abla} shows that both mix-up methods can help feature alignment with more distinct features but the hybrid strategy raises the best performance. More ablations are in the Appendix. Moreover, since masks generated by SAM do not contain semantic labels, we conduct an additional experiment to prove the validity of instance-level augmentation by replacing the generated SAM instance masks with the ground truth semantic masks under the setting (6). Compared with the original result (mIoU=44.0), the performance of using ground truth semantic masks is mIoU=42.7, demonstrating that although masks generated by SAM cannot be identical to the ground truth, the contained semantic information is consistent, and the randomness of our augmentation further improve the performance.

\begin{wraptable}{r}{0.5\textwidth}
\vspace{-2ex}
\centering
\caption{The effect of image-point coverage. Different numbers represent the number of used pictures of nuScenes.}
\resizebox{\linewidth}{!}{
\begin{tabular}{c|c|c|c|c}
\toprule
Covered Images for Alignment             & baseline    & 2 & 4 & 6    \\ \midrule
nuScenes $\rightarrow$ Sem.KITTI & 27.7 & 43.9  & 45.6 & \textbf{48.5} \\ \midrule
Sem.KITTI $\rightarrow$ nuScenes & 28.1 & 38.4  & 40.6 & \textbf{42.9} \\ \bottomrule
\end{tabular}
}
\label{tab_cover}

\vspace{-2ex}
\end{wraptable}


\noindent \textbf{Effect of Different Point-to-Pixel Coverage for Alignment.} For point clouds not covered by images, we do not calculate feature alignment loss and solely calculate segmentation loss with ground truth or pseudo label. When conducting the experiments, we used all available images to ensure that as many point cloud data as possible could find corresponding features on the pictures for SAM-guided feature alignment. We also conduct additional experiments to demonstrate the impact of the number of available images for feature alignment, as shown in Tab.\ref{tab_cover}. The results indicate that a greater number of available images correlates with improved experimental outcomes. Importantly, as long as SAM-guided feature alignment is achievable, the performance does not significantly degrade even with limited coverage.

\noindent \textbf{Effect of Vision Foundation Model} We also extend our method to other vision foundation models, such as InternImage~\cite{wang2023internimage}, serving for image-based tasks. Specifically, we replace the SAM-based image encoder with InternImage to guide the feature alignment in a similar manner. Compared with the baseline (mIoU=27.7), the performance of using InternImage is mIoU=36.9. A consistent performance gain can be obtained, which also verifies and validates our insight, \ie, the generic feature space of the VFM can ease the feature alignment.



    


\subsection{More Challenging Tasks}
Since we achieve the purpose of mapping data from different domains into a unified feature space, the extracted feature can be used for some more challenging tasks. We show some extension results of our method in Tab.~\ref{tab_extension}. The left subtable shows the results of UDA for panoptic segmentation, a more challenging task requiring instance-level predictions. With more accurate and fine-grained semantic prediction, our method achieves promising results. The right subtable shows the results of domain generalization, in which target data only can be used for testing. In this subtable, models are trained with nuScenes and SemanticKITTI and then evaluated with A2D2 dataset. With the ability of stronger data-to-feature mapping, our method outperforms the current SOTA method~\cite{li2023bev}. In the future, we seek to explore the potential of our method on more tasks, such as 3D detection.

\begin{table}[]
\centering

\caption{Extension on more challenging tasks, such as UDA for Panoptic Segmentation(left) and Domain Generalization(right), where N, S, A represent nuScenes, SemanticKITTI and A2D2 dataset.}
    \vspace{-3mm}
\begin{subtable}{0.7\linewidth}

\resizebox{\linewidth}{!}{
\begin{tabular}{c|l|c|ccc|c}
\toprule
\textbf{Task} & \textbf{Method} & \textbf{PQ} & \textbf{PQ}$^\dag$ & \textbf{RQ} & \textbf{SQ} & \textbf{mIoU}
\\\midrule
\multirow{4}{*}{nuScenes $\rightarrow$ Sem.KITTI} 
& Source only & $14.0$  & $21.6$  & $19.9$  & $55.8$  &  $27.7$     \\
& PL     & $15.9$  & $22.7$  & $22.2$  & $58.1$  &  $29.7$     \\
& \textbf{Ours}   & $\mathbf{34.3}$  & $\mathbf{38.4}$  & $\mathbf{42.6}$  & $\mathbf{55.9}$  &  $\mathbf{48.5}$     \\ \cmidrule{2-7}
& Oracle & $50.5$  & $52.2$  & $57.8$  & $77.2$  &  $70.3$     \\
\midrule
\multirow{4}{*}{Sem.KITTI $\rightarrow$ nuScenes} 
& Source only & $15.6$  & $22.1$  & $20.7$  & $52.7$  &  $28.2$   \\
& PL     & $16.8$  & $23.0$  & $21.7$  & $48.3$  &  $29.0$   \\
& \textbf{Ours}   & $\mathbf{24.6}$  & $\mathbf{30.7}$  & $\mathbf{30.8}$  & $\mathbf{60.0}$  &  $\mathbf{42.9}$   \\ \cmidrule{2-7}
& Oracle & $40.7$  & $44.9$  & $47.2$  & $83.8$  &  $78.3$   \\
\bottomrule
\end{tabular}
}

\end{subtable}
\begin{subtable}[t]{0.25\linewidth}

\resizebox{\linewidth}{!}{
\begin{tabular}{cc}
\toprule
 \textbf{Method}        & \textbf{N,S} $\rightarrow$ \textbf{A} \\ \midrule
Baseline & $45.0$                \\ \midrule
\makecell[c]{xMUDA \\ \cite{jaritz2022cross}}    & $44.9$                \\
\makecell[c]{Dual-Cross \\ \cite{li2022cross}}    & $41.3$                \\
\makecell[c]{BEV-DG \\ \cite{li2023bev}}   & $55.1$                \\ \midrule
\textbf{Ours}     & $\mathbf{57.2}$        \\
\bottomrule       
\end{tabular}
}

\end{subtable}
\label{tab_extension}
\vspace{-5mm}
\end{table}

\section{Conclusion}
In this paper, we acknowledge the limitations of existing UDA methods in handling the domain discrepancy present in 3D point cloud data and propose a novel paradigm to unify feature representations across diverse 3D domains by leveraging the powerful generalization capabilities of the vision foundation model, significantly enhancing the adaptability of 3D segmentation models. 
Hybrid feature augmentation strategy is also proposed for better 3D-SAM feature alignment. 
Extensive experiments show that our method surpasses all compared SOTA methods by a large margin. 


%
%
\bibliographystyle{splncs04}
\bibliography{main}


\end{document}


\title{Learning to Adapt SAM for Segmenting Cross-domain Point Clouds} 


\author{\large{Supplementary Materia}}

\authorrunning{Peng. et al.}


\institute{}

\maketitle

In this appendix section, we provide more supplementary material to support the findings and observations drawn in the main body of this paper.


\begin{figure*}[h]
    \centering
    \includegraphics[width=1.0\columnwidth]{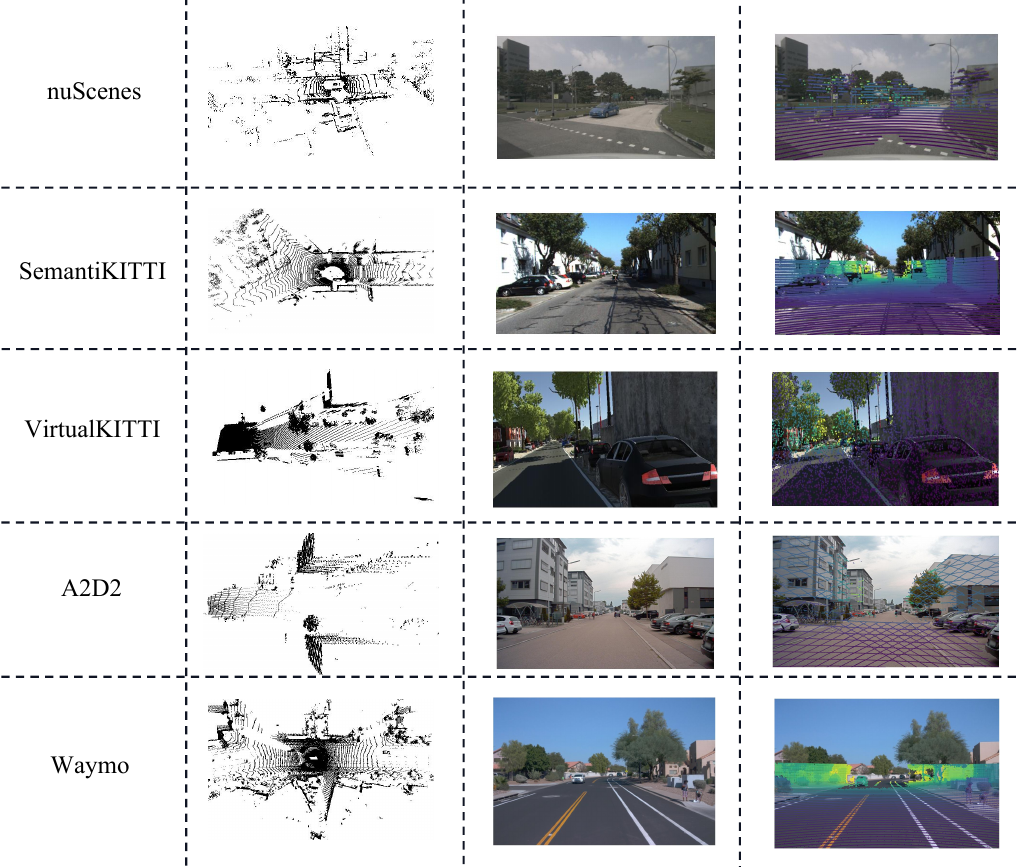}
    \caption{Visualization of each used dataset. From left to right, the figure shows the point cloud, one image corresponding to the point cloud,  and the projection of the point cloud on the image.}
    \label{fig_dataset_sup}
\end{figure*}

\section*{A. Dataset Details}
\label{sec_dataset_sup}
We conduct extensive experiments under several domain shift scenarios with five well-known large-scale datasets, including nuScenes, SemanticKITTI, VirtualKITTI, A2D2, and Waymo. All of these datasets provide point clouds and corresponding images captured by distinct devices resulting in different data representations, thus we will give more detailed information about them. (1) nuScenes contains 1000 driving scenes with 20 seconds for each scene, taken at 2Hz. The scenes are split into train (28,130 keyframes), validation (6,019 keyframes), and corresponding point-wise 3D semantic labels provided by nuScenes-Lidarseg. (2) SemanticKITTI features a large-angle front camera and a 64-layer LiDAR and the captured data from scenes 0, 1, 2, 3, 4, 5, 6, 7, 9, and 10 are used for training while scenes 8 as validation at most experiments. (3) VirtualKITTI consists of 5 driving scenes created with the Unity game engine. VirtualKITTI does not simulate LiDAR but rather provides a dense depth map with semantic labels, so we use the 2D-to-3D projecting to generate a point cloud from the depth map. (4) A2D2 consists of 20 drives corresponding to 28,637 frames. The point cloud comes from three 16-layer front LiDARs (left, center, right), and the semantic labeling was carried out in the 2D image. (5) Waymo offers 2,860 temporal sequences captured by five cameras and one LiDAR in three different geographical locations, leading to a total of 100k labeled data, making it larger than existing datasets that offer point-wise segmentation labels. We visualize these five datasets with one sample to show the data domain difference in Fig.~\ref{fig_dataset_sup}.


\begin{table}[h]
\centering
\caption{Effect of different point mix-up strategies for the hybrid feature augmentation.}
\setlength\tabcolsep{5.0pt}
\vspace{-2mm}
\scriptsize
\begin{tabular}{cccc|c}
\toprule
 \multicolumn{3}{c}{\makecell[c]{Scene-level Feature Augmentation \\ \midrule}} & \multirow{2}{*}{\makecell[c]{Instance-level \\ Feature Augmentation}} & \multirow{2}{*}{mIoU} \\
 Polar-based      & Range-based     & Laser-based     &                                                      &                       \\
\midrule
 $\checkmark$ &  &  &       &           37.0   \\
              & $\checkmark$ &   &                &   38.5   \\
              &  & $\checkmark$  &     &             37.6   \\ \midrule
              &  &  &  $\checkmark$    &             40.1   \\ 
 $\checkmark$ & $\checkmark$ & $\checkmark$ &                  &      39.0   \\
 $\checkmark$ & $\checkmark$ & $\checkmark$ &  $\checkmark$    &     \textbf{44.0}   \\
\bottomrule
\end{tabular}
\label{tab_mixup_sup}
\end{table}
\vspace{-2mm}

\section*{B. More Experimental Data}
\label{sec_exp_sup}

We conduct further ablation on the same setting in the Ablation Study for hybrid feature augmentation to thoroughly analyze its effect. We take all the point mix-up strategies used in it apart to conduct separate experiments as shown in Tab.~\ref{tab_mixup_sup} and these results show that all of them contribute to our method. So we combine them with random selection to form the hybrid feature augmentation. This comprehensive approach boosts the performance since it can provide different data with distinct feature patterns as much as possible for SAM-guided feature alignment. Additionally, we found that this approach can be used in other multi-modal methods as a normal data augmentation method. Taking MM2D3D as an example, we add this module into it and Tab.~\ref{tab_aug_sup} shows that our feature augmentation approach also improves its performance. However, this feature augmentation approach contributes to our method in a better way because of our effective SAM-guided feature alignment. 

\begin{table}[h]
\centering
\caption{Generalization of feature augmentation for other multi-modal UDA methods.}
\vspace{-2mm}
\tiny
\begin{tabular}{c|c|c|c|c}
\toprule
Source only & \makecell[c]{MM2D3D } & \makecell[c]{MM2D3D \\ w/ feature augmentation} & \makecell[c]{Ours \\ w/o feature augmentation} & Ours \\ \midrule
27.7  &  30.4  &  33.2   &  34.0   &   48.5      \\
\bottomrule
\end{tabular}
\label{tab_aug_sup}
\vspace{-4mm}
\end{table}

\section*{C. More Qualitative Results}
\label{sec_res_sup}
\begin{figure*}
    \centering
    \includegraphics[width=1.0\columnwidth]{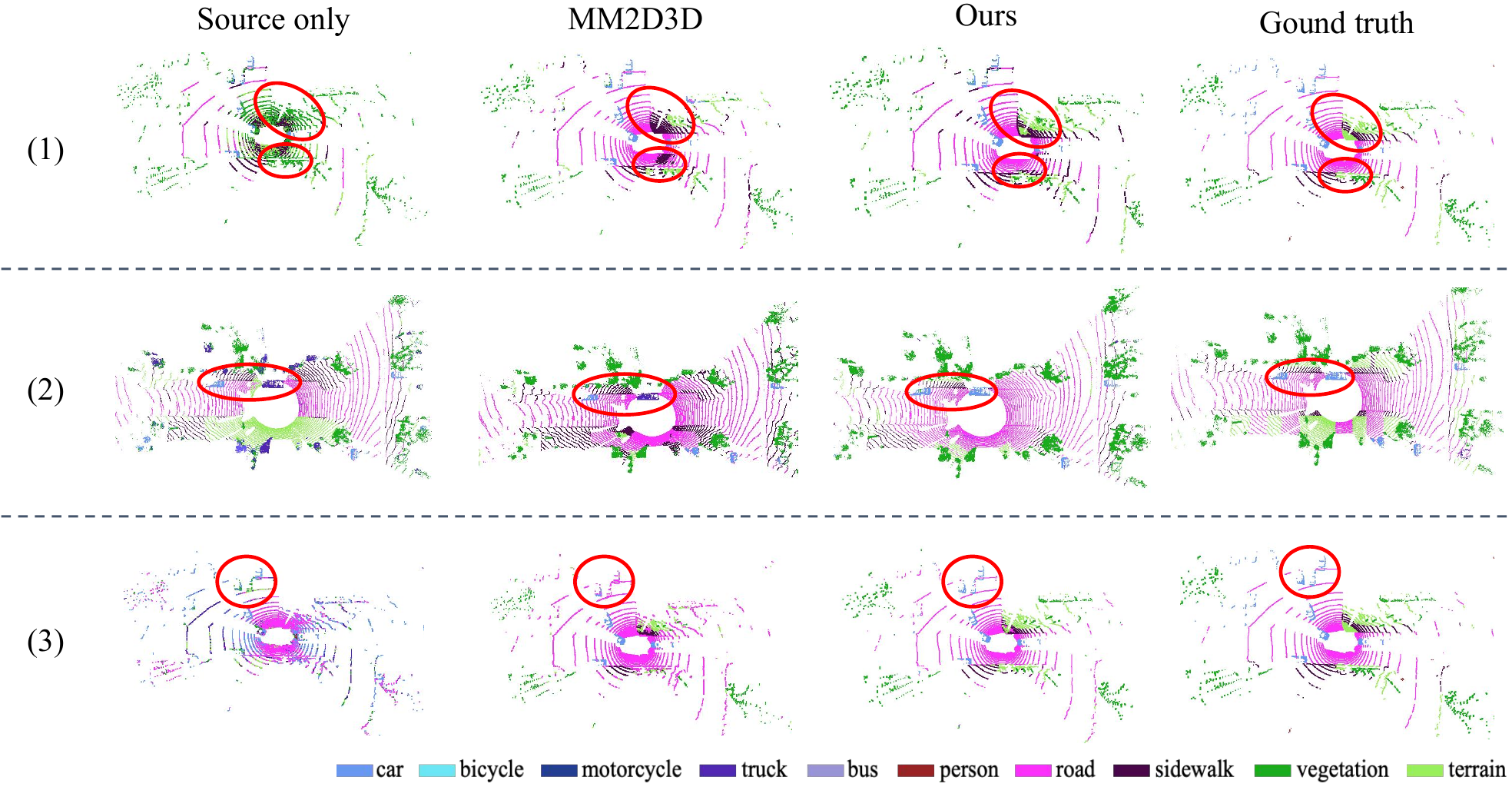}
    \caption{Visualization results of the domain adaptation from more domain shift scenarios, including (1)SemanticKITII-to-nuScenes, (2) nuScenes-to-Waymo,  and (3) Waymo-to-nuScenes.}
    \label{fig_res_sup}
     \vspace{-3mm}
\end{figure*}

Except for the visualization results of the domain adaptation from nuScenes to SemanticKITII shown in the main paper, we provide additional qualitative results in Fig.~\ref{fig_res_sup} representing other introduced domain shift scenarios in the experiment including SemanticKITII-to-nuScenes, nuScenes-to-Waymo, and Waymo-to-nuScenes. Our method has more accurate predictions for the cars, sidewalks, terrain, etc.

